\documentclass{article} 
\usepackage{iclr2027_conference,times}

\usepackage{amsmath,amsfonts,bm}

\def\eqref#1{equation~\ref{#1}}

\def\1{\bm{1}}

\DeclareMathAlphabet{\mathsfit}{\encodingdefault}{\sfdefault}{m}{sl}
\SetMathAlphabet{\mathsfit}{bold}{\encodingdefault}{\sfdefault}{bx}{n}

\usepackage{hyperref}
\usepackage{url}

\usepackage{booktabs}
\usepackage{graphicx}
\usepackage{multirow}
\usepackage{makecell}
\usepackage{tabularx}

\usepackage{wrapfig}

\usepackage{cleveref}

\usepackage[table]{xcolor}
\definecolor{EchoRowBlue}{RGB}{225,239,255}

\title{Echo: Coupled Action and Task Status Signal Denoising for Vision Language Action Models}

\author{
{\small
\begin{tabular}{@{}l@{}}
\bfseries
Yufeng Liu$^{1,2,*}$,
Bohao Fan$^{2,*}$,
Juntu Zhao$^{1,2}$,
Xinyi Wang$^{2,4}$,
Hang Yu$^{2,5}$,
Wenzhuan Luo$^{2}$ \\[0pt]
\bfseries
Yingzhi Wang$^{2,6}$,
Qinghai Li$^{2}$,
Yingdong Hu$^{2}$,
Junliang Guo$^{2}$,
Richard Wang$^{2}$ \\[0pt]
\bfseries
Dequan Wang$^{1,\dagger}$,
Yang Gao$^{2,3,\dagger}$,
Cheng Ma$^{2,\ddagger}$
\end{tabular}
}\\[12pt]
{\small\normalfont
$^{1}$SJTU
\hspace{0.3em}
$^{2}$Spirit AI
\hspace{0.3em}
$^{3}$THU
\hspace{0.3em}
$^{4}$USTC
\hspace{0.3em}
$^{5}$Tongji
\hspace{0.3em}
$^{6}$HIT
}
}

\iclrfinalcopy 
\begin{document}

\maketitle

\vspace{-26pt}
{\raggedright\normalsize\hypersetup{hidelinks}\color[RGB]{102,170,122}\urlstyle{same}
\noindent\hspace*{\tabcolsep}Project page: \url{https://lyfeng001.github.io/Echo/}\par
}
\vspace{5pt}

\makeatletter
\begingroup
\renewcommand\@makefntext[1]{#1}
\footnotetext{%
\hspace*{1em}%
\begin{tabular}[t]{@{}r@{\hspace{0em}}l@{}}
$^{\dagger}$ & Corresponding authors.
               $^{\ddagger}$ Project leader. \\[0pt]
$^{*}$       & Equal contribution. 
\end{tabular}%
}
\endgroup
\makeatother

\begin{abstract}
    Recognizing task completion and failure enables robots to stop upon success and initiate recovery or replanning after failure. Existing methods augment Vision Language Action (VLA) policies with task status signal prediction to support these execution decisions. However, learning and integrating these signals can require fine-grained supervision, external monitoring models, or complex architectural designs. Moreover, the potential of task status signals to directly guide action generation remains underexplored. We present Echo, a framework that combines task status signal learning without frame-level progress annotations with a compact design for status prediction and action generation within a single VLA. Specifically, Echo represents task status with stop and failure signals, added as two dimensions of each action chunk and denoised together with robot actions. Their combination distinguishes ongoing execution, successful completion, and failure. To learn these signals without fine-grained progress annotations, Echo constructs successful, failed, and recovery trajectories with complementary supervision properties. To enable status signal-guided action generation, Echo controls interactions between the signals and actions and denoises the signals faster than the actions. These allow later action denoising steps to condition on cleaner status estimates while preserving accurate signal prediction. Extensive real-world manipulation experiments show that Echo accurately predicts task status and consistently improves policy performance.
\end{abstract}

\section{Introduction}

\begin{wrapfigure}{r}{0.47\textwidth}
    \vspace{-\intextsep}
    \centering
    \includegraphics[width=\linewidth]{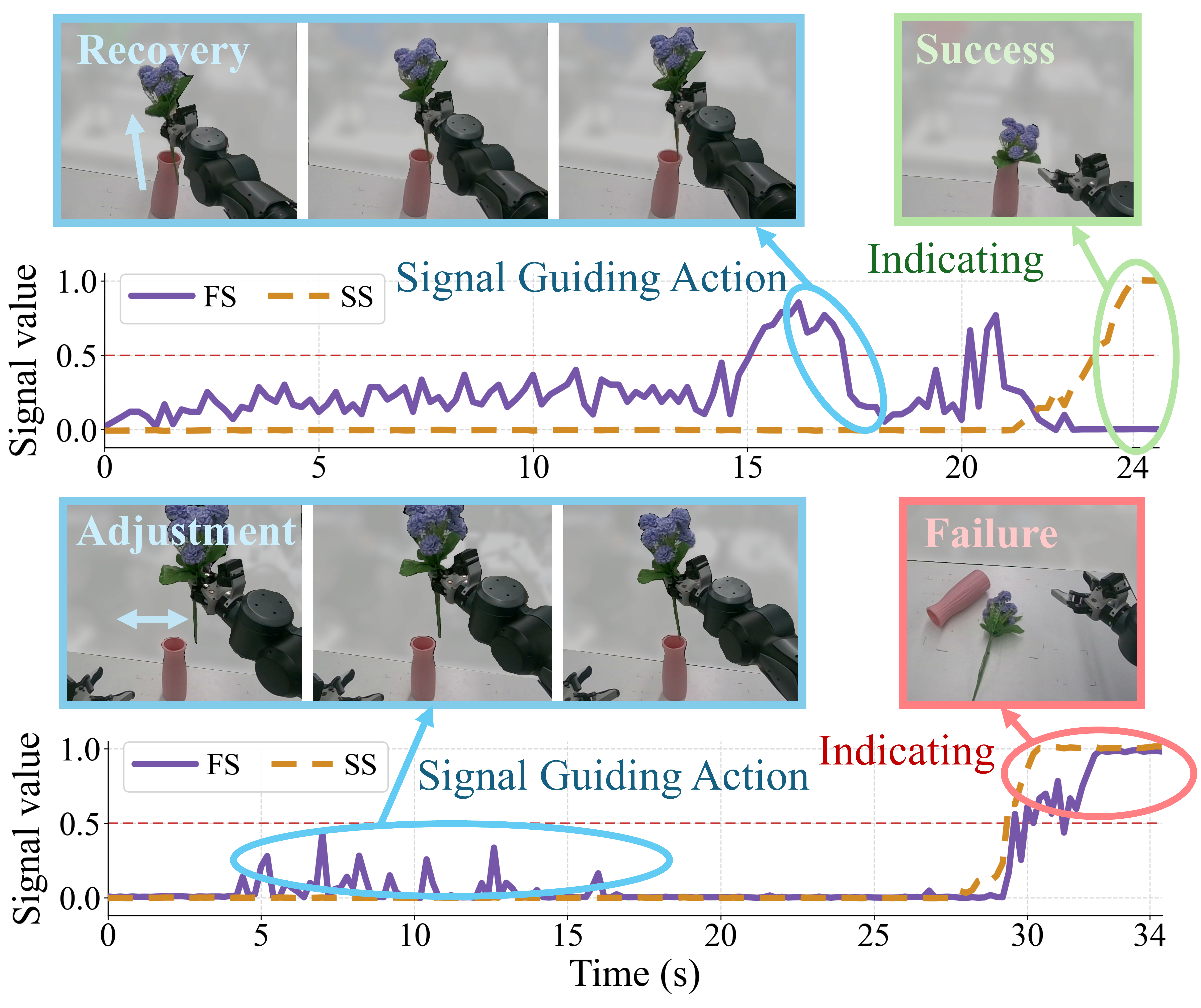}
    \caption{A low stop signal (SS, orange) value indicates ongoing execution, where a high failure signal (FS, purple) value marks recoverable failure and guides correction. High SS with low or high FS values indicates success (top) or terminal failure (bottom) respectively.}
    \label{fig:teaser}
    \vspace{-\intextsep}
    \vspace{-\intextsep}
    \vspace{-\intextsep}
    \vspace{-\intextsep}
    
\end{wrapfigure}

Vision Language Action (VLA) models have shown strong capabilities in robotic manipulation by mapping visual observations and language instructions to robot actions~\citep{intelligence2025pi_,bjorck2025gr00t,team2026xiaomi}. Reliable execution also requires recognizing whether a task is ongoing, has succeeded, or has failed to determine when to continue, stop, or intervene. Existing methods predict progress, completion, or failure signals either within VLA policies or through external monitoring models~\citep{apanasevich2026green,feng2026procvlm}.

However, existing approaches face two limitations. First, learning and integrating these signals can require fine-grained frame-level supervision, external monitoring models, or complex architectural designs~\citep{pacaud2025scaling,feng2026procvlm,yan2026progressvla,deng2026uniintervene}. Second, the predicted signals are typically used only for monitoring, leaving their potential to improve action generation underexplored~\citep{pacaud2025scaling,huang2026stageguard}.


We present Echo, a compact framework that jointly denoises robot actions and task status signals within a single VLA, learning the signals without frame-level progress annotations and using them to improve action generation. Echo represents task status through a stop signal (SS), indicating whether execution should terminate, and a failure signal (FS), indicating whether execution is in a failure state. Low SS indicates ongoing execution, while high SS with low FS indicates successful completion, and high SS with high FS indicates terminal failure. Within ongoing execution, elevated FS indicates a recoverable failure, as shown in Figure~\ref{fig:teaser}. Echo appends SS and FS as two additional dimensions of each action chunk and denoises them together with robot actions. This formulation reuses the VLA’s visual, language, and robot-state features with little additional parameter overhead.

To learn these signals without frame-level progress annotations, Echo combines successful demonstrations, failed policy rollouts, pre-intervention failure segments, and recovery demonstrations using subtask-level prompts together with success or failure labels. Failed rollouts and pre-intervention failure segments supervise terminal and recoverable failures, respectively, while their actions are excluded from imitation. Recovery demonstrations pair similar status supervision with corrective actions at corresponding recoverable failure states, linking failure recognition to recovery behavior.

To use the learned status information during action generation, Echo controls interactions among SS, FS, and robot actions while preserving accurate and complementary signal predictions. It also introduces asynchronous denoising, denoising signal tokens faster than action tokens so that later action denoising steps condition on cleaner status estimates. When the policy recognizes a recoverable failure with low SS and elevated FS, these estimates, together with the current observations, can guide action generation toward corrective behavior learned from recovery demonstrations.

Experiments on five real-world manipulation tasks demonstrate accurate task-status prediction and improved performance over action-only baselines, with larger out-of-distribution gains. Component ablations, signal interventions, and failure-data scaling analyses examine how status signals influence action generation and how different data sources contribute to prediction and execution.

Our contributions are threefold: 1) We introduce a compact task status representation that incorporates stop and failure signals into VLA action chunks with little additional parameter overhead. 2) We develop a supervision scheme that learns these signals from successful, failed, and recovery data without frame-level progress annotations. 3) We design controlled interactions and asynchronous denoising to couple status prediction with action generation.
\section{Related Works}

\textbf{Task Status Signal Modeling.}\enspace
Task status signals characterize task progress, completion, and failure~\citep{kim2026progvla,liu2026palm,dai2026see,liu2026self,liufu2026repo,zhao2026flare}. Visual-language reward models learn execution quality through representation learning, temporal supervision, and trajectory comparisons~\citep{ma2023liv,alakuijala2024video,ayalew2025progressor,zhang2025rewind,lee2026roboreward,liang2026robometer}. When deployed as standalone evaluators, these models introduce additional evaluation alongside the control policy. Complementary approaches predict completion~\citep{yang2025seqvla} or detect failures using policy features, temporal consistency, uncertainty, and visual reasoning~\citep{gu2026safe,park2026hide,agia2024unpacking,xu2025can,duan2024aha}. Progress can also be extracted through lightweight probes~\citep{bhardwaj2026decoding}. Although feature-based monitoring reduces overhead, accurate signal extraction alone does not ensure action feedback. Jointly achieving reliable task status prediction and signal-guided action adjustment with low overhead remains underexplored.

\textbf{Task Status Signals for Action Improvement.}\enspace
Task status feedback improves policies through several mechanisms. VLAC~\citep{zhai2025vision} uses critic tokens to estimate rewards for  
online reinforcement learning, while GRAPE~\citep{zhang2024grape} optimizes policies using trajectory preferences; both require additional policy updates. Other methods use execution assessments to trigger backtracking, intervention, or corrective control~\citep{ma2026cyclevla,deng2026uniintervene,yang2026fpc,lin2025failsafe,li2026sentinel,liu2024model}. More directly, VLA-Corrector~\citep{pan2026vla} monitors visual dynamics deviations and applies gradient-guided corrective replanning, while ProgressVLA~\citep{yan2026progressvla} propagates progress gradients through a world model to guide action denoising. These mechanisms demonstrate the utility of execution feedback, but introduce additional policy optimization, monitoring and recovery components, or inference-time gradient computation. Preserving accurate status estimates while directly improving action generation with minimal additional training and inference overhead remains an open challenge.

\section{Methodology}
\label{sec:method}

\begin{figure*}[t]
    \centering
    \includegraphics[width=1\linewidth]{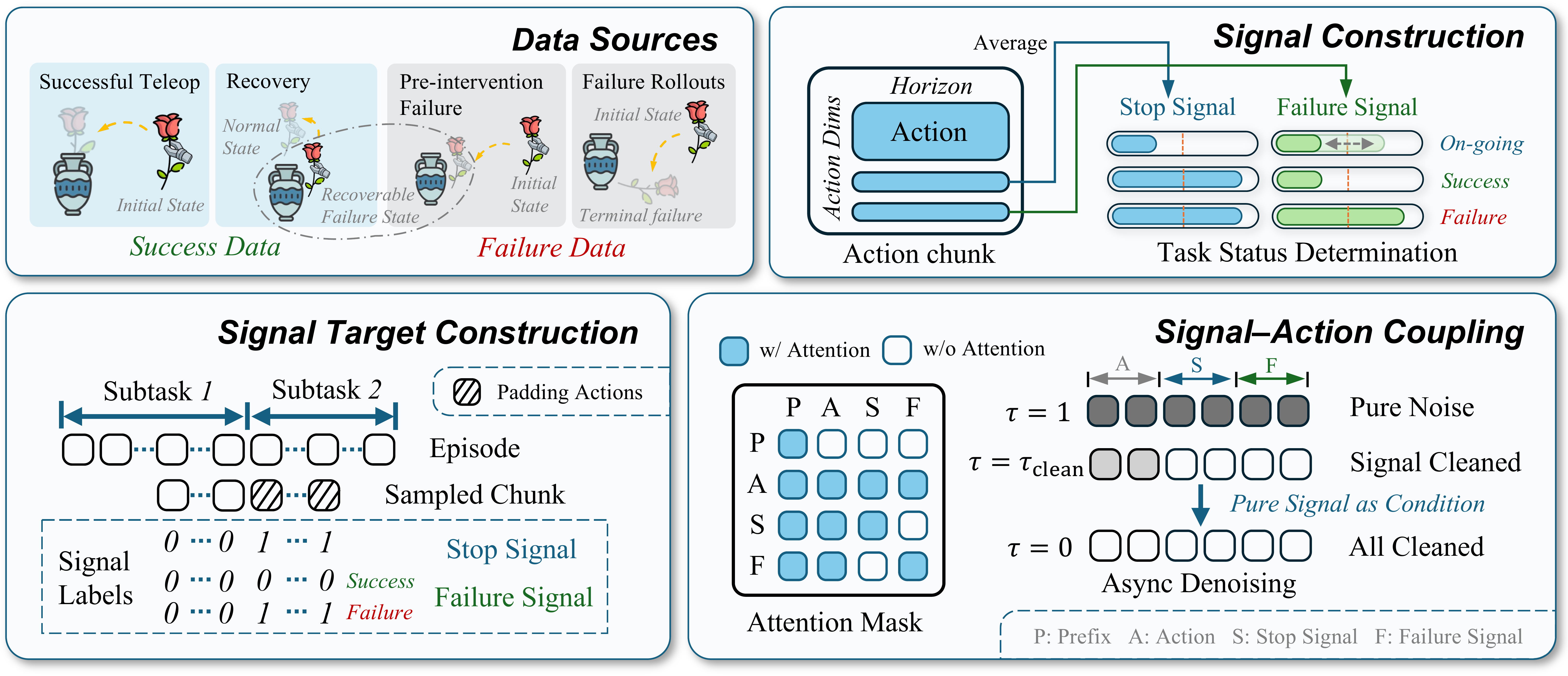}
    \caption{Overview of Echo. Four data types provide complementary action and signal supervision, with failed actions excluded from imitation. Stop (SS) and failure (FS) signals are appended to action chunks, with padding-based target construction illustrated on the left. The attention mask couples actions with both signals while blocking direct SS--FS interactions. Asynchronous denoising provides cleaner signals to guide subsequent action updates.}
    \label{fig:overview}
\end{figure*}


\subsection{Preliminaries}
\label{sec:preliminaries}

\paragraph{Action chunk representation and padding.}
An episode contains either a single task or multiple consecutive subtasks. For multi-subtask episodes, we manually annotate subtask boundaries and associate each segment with a language instruction. At episode timestep $t$, the policy takes the current observation, robot state, and instruction as input and predicts an action chunk $\mathbf{A}_t\in\mathbb{R}^{H\times d_a}$. Here, $t$ is the starting timestep of the chunk, $H$ is its prediction horizon, and $d_a$ is the action dimension. Each action $\mathbf{a}_{t,i}\in\mathbb{R}^{d_a}$ corresponds to timestep $t+i$ and is expressed relative to the robot state at the beginning of the chunk.

A sampled chunk may extend beyond the current segment, entering a subsequent subtask or exceeding the episode length. Let $T$ denote the segment's last valid action timestep. We identify positions beyond this boundary using a padding mask:
\begin{equation}
    p_{t,i}
    =
    \begin{cases}
        0, & t+i\leq T,\\
        1, & t+i>T,
    \end{cases}
    \qquad i=0,\ldots,H-1.
    \label{eq:padding_mask}
\end{equation}
We repeat the last valid action target at padded positions:
\begin{equation}
\tilde{\mathbf a}_{t,i}
=
(1-p_{t,i})\mathbf a_{t,i}
+
p_{t,i}\mathbf a_{t,T-t}.
\label{eq:action_padding}
\end{equation}
Besides excluding actions beyond the current segment, padding naturally provides boundary-based signal labels. For example, when a segment ends in task completion, its padding mask indicates where stopping should be predicted. Echo combines this mask with the segment type to construct task status signal targets without additional frame-level progress annotations.

\paragraph{Flow matching.}
Echo builds on a flow-based VLA policy. Given a clean sample $\mathbf{x}_0$ and Gaussian noise $\boldsymbol{\epsilon}\sim\mathcal{N}(\mathbf{0},\mathbf{I})$, we construct
\begin{equation}
    \mathbf{x}_{\tau}
    =
    (1-\tau)\mathbf{x}_0+\tau\boldsymbol{\epsilon},
    \qquad \tau\in[0,1],
    \label{eq:flow_path}
\end{equation}
where $\tau=0$ denotes clean data and $\tau=1$ denotes noise. The model predicts a velocity $\mathbf{v}_{\theta}(\mathbf{x}_{\tau},\tau,\mathbf{c})$, conditioned on the observations, robot state, and instruction, collectively denoted by $\mathbf{c}$. Training matches the predicted velocity to $\boldsymbol{\epsilon}-\mathbf{x}_0$. Inference integrates the predicted velocity from $\tau=1$ to $\tau=0$.

\subsection{Echo Framework}
\label{sec:echo}


\subsubsection{Task-Status Representation}
\label{sec:signal_modeling}

Echo uses two complementary signals as shown in Figure~\ref{fig:overview}: a stop signal (SS), indicating whether execution should terminate, and a failure signal (FS), indicating whether execution is in a failure state. Both signals share the action horizon and are represented as column vectors $\mathbf{S}_t,\mathbf{F}_t\in\mathbb{R}^{H\times 1}$:
\begin{equation}
    \mathbf{S}_t=[s_{t,0},\ldots,s_{t,H-1}]^{\top},
    \qquad
    \mathbf{F}_t=[f_{t,0},\ldots,f_{t,H-1}]^{\top}.
    \label{eq:status_vectors}
\end{equation}
Concatenating them with the action matrix along the feature dimension gives the joint target
\begin{equation}
    \mathbf{X}_t
    =
    [\tilde{\mathbf{A}}_t,\mathbf{S}_t,\mathbf{F}_t]
    \in\mathbb{R}^{H\times(d_a+2)}.
    \label{eq:joint_representation}
\end{equation}

We average each predicted signal within the current chunk to obtain scalar estimates for timestep $t$:
\begin{equation}
    \bar{s}_t
    =
    \frac{1}{H}\sum_{i=0}^{H-1}\hat{s}_{t,i},
    \qquad
    \bar{f}_t
    =
    \frac{1}{H}\sum_{i=0}^{H-1}\hat{f}_{t,i}.
    \label{eq:signal_average}
\end{equation}
Given thresholds $\delta_s$ and $\delta_f$, the execution status is
\begin{equation}
    z_t
    =
    \begin{cases}
        \mathrm{Ongoing},
        & \bar{s}_t<\delta_s,\\
        \mathrm{Success},
        & \bar{s}_t\geq\delta_s
          \land \bar{f}_t<\delta_f,\\
        \mathrm{Terminal\ failure},
        & \bar{s}_t\geq\delta_s
          \land \bar{f}_t\geq\delta_f.
    \end{cases}
    \label{eq:execution_status}
\end{equation}

\definecolor{EchoPadBlue}{HTML}{174A7E}
\definecolor{EchoExampleBg}{HTML}{F5F6F7}
\definecolor{EchoSuccessBg}{HTML}{EFF6FC}
\definecolor{EchoFailureBg}{HTML}{F3F3F3}

\begin{table}[t]
    
    \caption{Action and task-status supervision for four data types. $H$ is the chunk horizon, $n$ counts non-padding positions, and $n_t$ is defined in Eq.~\ref{eq:recovery_progress}. $\mathbf{0}_k$ and $\mathbf{1}_k$ denote $k$ zeros and ones. Blue text marks padding positions. ``All'', ``Non-pad'', and ``--'' denote supervision over the full chunk, non-padding positions only, and no supervision, respectively.}
    \label{tab:data_supervision}

    \begin{center}
    \setlength{\tabcolsep}{3pt}
    \setlength{\fboxsep}{1.5pt}
    \edef\EchoTableWidth{\the\linewidth}
    \renewcommand{\arraystretch}{1.1}
    \newcommand{\EchoPair}[2]{\begin{tabular}[c]{@{}c@{}}#1\\#2\end{tabular}}
    \small
    \begin{tabularx}{\linewidth}{
        >{\raggedright\arraybackslash}p{0.14\linewidth}
        >{\centering\arraybackslash}p{0.085\linewidth}
        >{\centering\arraybackslash}p{0.115\linewidth}
        >{\hsize=1.15\hsize\linewidth=\hsize\centering\arraybackslash}X
        >{\hsize=0.85\hsize\linewidth=\hsize\centering\arraybackslash}X
        >{\hsize=1.15\hsize\linewidth=\hsize\centering\arraybackslash}X
        >{\hsize=0.85\hsize\linewidth=\hsize\centering\arraybackslash}X
    }
        \toprule
        \multicolumn{7}{@{}c@{}}{\colorbox{EchoExampleBg}{\makebox[\dimexpr\EchoTableWidth-2\fboxsep\relax][c]{\strut\textbf{Sampled action chunks:}\quad \textit{No padding:} $[\mathbf{a}_1,\ldots,\mathbf{a}_H]$\qquad \textit{With padding:} $[\mathbf{a}_1,\ldots,\mathbf{a}_n,\textcolor{EchoPadBlue}{\text{pad},\ldots,\text{pad}}]$}}} \\
        \midrule
        \raisebox{-0.8\baselineskip}[0pt][0pt]{\textbf{Data type}}
        & \raisebox{-0.8\baselineskip}[0pt][0pt]{\textbf{Padding?}}
        & \raisebox{-0.8\baselineskip}[0pt][0pt]{\EchoPair{\textbf{Action}}{\textbf{Supervision}}}
        & \multicolumn{2}{c}{\textbf{Stop Signal (SS)}}
        & \multicolumn{2}{c}{\textbf{Failure Signal (FS)}} \\
        \cmidrule(lr){4-5}\cmidrule(lr){6-7}
        & & & \textbf{Target} & \textbf{Supervision} & \textbf{Target} & \textbf{Supervision} \\
        \midrule
        \rowcolor{EchoSuccessBg}
        Success
        & \EchoPair{No}{Yes}
        & \EchoPair{All}{All}
        & \EchoPair{$[\mathbf{0}_H]$}{$[\mathbf{0}_n,\textcolor{EchoPadBlue}{\mathbf{1}_{H-n}}]$}
        & \EchoPair{All}{All}
        & \EchoPair{$[\mathbf{0}_H]$}{$[\mathbf{0}_n,\textcolor{EchoPadBlue}{\mathbf{0}_{H-n}}]$}
        & \EchoPair{All}{All} \\
        \cmidrule{1-7}
        \rowcolor{EchoSuccessBg}
        Recovery
        & \EchoPair{No}{Yes}
        & \EchoPair{All}{Non-pad}
        & \EchoPair{$[\mathbf{0}_H]$}{$[\mathbf{0}_n,\textcolor{EchoPadBlue}{\mathbf{0}_{H-n}}]$}
        & \EchoPair{All}{Non-pad}
        & \EchoPair{$[\mathbf{0}_{n_t},\mathbf{1}_{H-n_t}]$}{$[\mathbf{0}_{n_t},\mathbf{1}_{H-n_t}]$}
        & \EchoPair{All}{All} \\
        \cmidrule{1-7}
        \rowcolor{EchoFailureBg}
        Failure
        & \EchoPair{No}{Yes}
        & \EchoPair{--}{--}
        & \EchoPair{$[\mathbf{0}_H]$}{$[\mathbf{0}_n,\textcolor{EchoPadBlue}{\mathbf{1}_{H-n}}]$}
        & \EchoPair{All}{All}
        & \EchoPair{--}{$[\mathbf{0}_n,\textcolor{EchoPadBlue}{\mathbf{1}_{H-n}}]$}
        & \EchoPair{--}{All} \\
        \cmidrule{1-7}
        \rowcolor{EchoFailureBg}
        \begin{tabular}[c]{@{}l@{}}Pre-intervention\\failure\end{tabular}
        & \EchoPair{No}{Yes}
        & \EchoPair{--}{--}
        & \EchoPair{$[\mathbf{0}_H]$}{$[\mathbf{0}_n,\textcolor{EchoPadBlue}{\mathbf{0}_{H-n}}]$}
        & \EchoPair{All}{All}
        & \EchoPair{--}{$[\mathbf{0}_n,\textcolor{EchoPadBlue}{\mathbf{1}_{H-n}}]$}
        & \EchoPair{--}{All} \\
        \bottomrule
    \end{tabularx}
    \end{center}
\end{table}

\subsubsection{Data Construction and Supervision}
\label{sec:data_supervision}

We collect successful demonstrations through teleoperation and terminal failure rollouts during policy testing. During policy execution, operators intervene when recoverable failures occur, yielding pre-intervention failure segments before human takeover and recovery demonstrations afterward. Recovery demonstrations may complete a local correction or the entire instructed task. The collected data are aggregated to update the policy for subsequent collection rounds.

We construct action, SS, and FS supervision from segment types, boundaries, and positions within recovery segments, without additional frame-level progress annotations. Table~\ref{tab:data_supervision} summarizes the targets and supervision regions. For a chunk with $n$ non-padding positions, the padding-mask target is displayed as $[\mathbf{0}_n,\mathbf{1}_{H-n}]$. Blue identifies padding positions, independently of their target values or whether they are supervised.

\textbf{Action supervision.}\enspace
Successful demonstrations and recovery demonstrations provide task-execution and corrective actions, respectively. We exclude actions from both terminal and pre-intervention failure data to avoid imitating unsuccessful behavior. Successful demonstrations receive action supervision over the entire chunk, including padding. Recovery actions are supervised only at non-padding positions: a recovery segment may end before the instructed task is complete, and learning its padded actions could cause the policy to stop prematurely.

\textbf{Stop signal supervision.}\enspace
Chunks without padding receive zero SS targets at all positions. For chunks containing padding, the targets depend on whether the segment boundary marks the end of execution. Successful demonstrations end upon completion, while terminal failure rollouts end after an unrecoverable failure. Both therefore use the padding mask as the SS target, with supervision over the entire chunk.

Pre-intervention segments end at operator takeover, and task execution continues through recovery. Their SS targets are therefore zero everywhere, including padding, and are fully supervised. For recovery segments, we also set the full SS target to zero but supervise only non-padding positions. Recovery may end after a local correction or upon full task completion, so its boundary does not consistently indicate whether execution should stop. Excluding padded positions avoids assigning an incorrect termination target.

\textbf{Failure signal supervision.}\enspace
Successful demonstrations contain successful execution without failure, so FS targets are zero over the entire chunk. For terminal and pre-intervention failure data, the segment-level labels indicate failure but do not identify its onset. We therefore omit FS supervision for chunks without padding. The segment endpoint, however, is known to be in a failure state. For chunks containing padding, we use the padding mask as the FS target and supervise the full chunk, assigning one to padded positions and zero elsewhere. This provides a boundary-based target without requiring the precise failure onset to be annotated.

Recovery demonstrations describe a transition from a recoverable failure state to a state from which successful task execution can resume. We encode this transition using binary chunk targets whose mean decreases with recovery progress. For a recovery segment spanning frames $t_0$ to $t_1$, with $t_1>t_0$, define
\begin{equation}
    u_t=\frac{t-t_0}{t_1-t_0},
    \qquad
    n_t=\operatorname{round}(H u_t).
    \label{eq:recovery_progress}
\end{equation}
Here, round rounds to the nearest integer, with ties to even. The FS target is $\mathbf{r}_t=[r_{t,0},\ldots,r_{t,H-1}]^{\top}$, where
\begin{equation}
    r_{t,i}
    =
    \begin{cases}
        0, & i<n_t,\\
        1, & i\geq n_t,
    \end{cases}
    \qquad i=0,\ldots,H-1.
    \label{eq:recovery_ramp}
\end{equation}
The first frame's chunk is all ones, and the last frame's chunk is all zeros. Intermediate chunks contain $n_t$ leading zeros followed by $H-n_t$ ones, shown as $[\mathbf{0}_{n_t},\mathbf{1}_{H-n_t}]$ in Table~\ref{tab:data_supervision}. The mean target, $1-n_t/H$, decreases approximately linearly with recovery progress. Unlike the padding boundary $n$, $n_t$ is determined by the starting frame's position within the recovery segment. FS is supervised over the full chunk, including padding.

\subsubsection{Status-Guided Action Generation}
\label{sec:signal_action_coupling}

We next use the learned status signals to guide action generation, including corrective behavior during ongoing execution. To this end, Echo combines controlled signal--action interactions with asynchronous denoising, allowing later action updates to condition on cleaner status estimates.

\textbf{Joint token processing.}\enspace
We project actions, SS, and FS into separate token groups and concatenate them with the multimodal prefix:
\begin{equation}
    \mathbf{H}
    =
    [\mathbf{H}^{P};\mathbf{H}^{A};\mathbf{H}^{S};\mathbf{H}^{F}],
    \qquad
    \mathbf{H}^{q}
    =
    \phi_q(\mathbf{x}^{q}_{\tau_q}),
    \quad q\in\{A,S,F\},
    \label{eq:token_groups}
\end{equation}
where $\mathbf{H}^{P}$ denotes the multimodal prefix tokens carrying the conditioning context, $\phi_q$ is the input projection for group $q$, and $\tau_q$ is its flow time. The concatenated tokens are processed together by a shared Transformer architecture. Action, SS, and FS tokens use separate MLPs for their feature transformations and separate output projections for velocity prediction.

To support different noise levels within the same sequence, we extend the flow-time input from one value per sample to one value per generated token. For batch size $B$ and concatenated generated-token length $L$, the time tensor has shape $B\times L$~\citep{black2025training}. Tokens belonging to the same group receive that group's flow time. The resulting time embeddings are passed to AdaRMS within the shared Transformer.

\textbf{Controlled signal-action interactions.}\enspace
Action tokens attend to the multimodal prefix and both signal groups, while SS and FS tokens attend to action tokens. These bidirectional interactions allow status estimates to guide action generation and action features to inform status prediction. We block direct SS--FS attention in both directions to reduce cross-signal interference and preserve their distinct termination and failure semantics.

\begin{wrapfigure}{r}{0.4\textwidth}
    \vspace{-\intextsep}
    \centering
    \includegraphics[width=\linewidth]{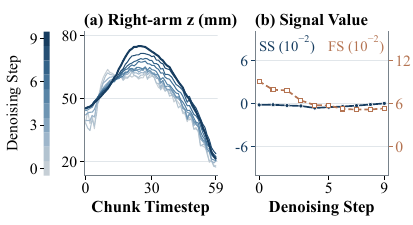}
    \caption{Intermediate clean-sample estimates under synchronous denoising. (a) Right-arm vertical action estimates; darker curves indicate later denoising steps. (b) Stop (SS) and failure (FS) signal estimates, plotted on separate scales. Both signals converge earlier than actions, with SS already near its zero target at the earliest denoising step. This motivates resolving signals earlier to guide subsequent action updates.}
    \label{fig:speed}
\end{wrapfigure}
\textbf{Motivating experiment: earlier status estimates.}
To examine when status information becomes available, we train a baseline that denoises signals and actions synchronously. At an intermediate flow time, its predicted velocity gives a clean-sample estimate:
\begin{equation}
    \hat{\mathbf{x}}_0(\tau)
    =
    \mathbf{x}_{\tau}
    -
    \tau\mathbf{v}_{\theta}(\mathbf{x}_{\tau},\tau,\mathbf{c}).
    \label{eq:clean_estimation}
\end{equation}

As shown in Figure~\ref{fig:speed}(b), both SS and FS approach their final estimates early in denoising. 
Action estimates, however, continue to change even between steps 8 and 9, the final two plotted steps (Figure~\ref{fig:speed}(a)). These observations motivate resolving signals earlier so that cleaner status estimates can guide the remaining action updates. Details are provided in Appendix~\ref{app:motivating_experiment}.

\textbf{Asynchronous denoising.}\enspace
Let $\tau_A$ be the action flow time. For signal group $q\in\{S,F\}$, we use
\begin{equation}
    \tau_q
    =
    \max\left(
        0,
        \frac{\tau_A-\tau_{\text{clean}}}{1-\tau_{\text{clean}}}
    \right),
    \qquad 0\leq\tau_{\text{clean}}<1,
    \label{eq:async_schedule}
\end{equation}
where $\tau_{\text{clean}}$ is the action flow time at which signal $q$ reaches its clean endpoint. Setting $\tau_{\text{clean}}=0$ gives synchronous denoising. With $\tau_{\text{clean}}>0$, the signal becomes clean before action denoising is complete.

During training, we sample $\tau_A$, obtain the signal times from Eq.~\ref{eq:async_schedule}, and perturb each group.
\begin{equation}
    \mathcal{L}
    =
    \mathbb{E}\left[
        \sum_{q\in\{A,S,F\}}
        \lambda_q
        \left\|
            \mathbf{M}^{q}\odot
            \left(
                \hat{\mathbf{v}}^{q}_{\theta}
                -(\boldsymbol{\epsilon}^{q}-\mathbf{x}^{q}_0)
            \right)
        \right\|_F^2
    \right].
    \label{eq:echo_objective}
\end{equation}
Here, $\lambda_A=1$, while $\lambda_S$ and $\lambda_F$ weight signal losses for chunks without padding to balance the numbers of padded and unpadded samples; both are set to $1$ for chunks containing padding. The binary masks $\mathbf{M}^{q}$ select the supervised positions in Table~\ref{tab:data_supervision} and are broadcast over feature dimensions.

At inference, each group starts from Gaussian noise and is updated according to its own flow-time schedule. Once a signal reaches its clean endpoint, its value is held fixed while action denoising continues. Through attention, later action updates can therefore use the cleaner status estimates. The final signal means provide the scalar predictions used for execution decisions.

\section{Experiments}
\label{sec:experiments}

We first report main results and comparisons with existing methods to evaluate the accuracy of Echo's task status signals and their benefit to task execution. We then analyze in-distribution (ID) and out-of-distribution (OOD) performance and ablate the model components. Finally, we intervene on the signals to verify their effect on action generation and vary the amount of failure data to examine the role of data sources.

\subsection{Experimental Setup}
\label{sec:exp_setup}

\begin{wrapfigure}{r}{0.41\textwidth}
    \vspace{-\intextsep}
    \centering
    \includegraphics[width=\linewidth]{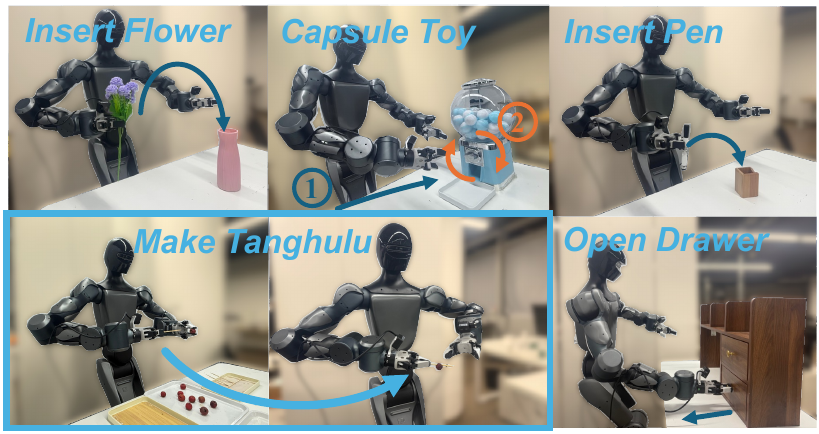}
    \caption{Real-world experimental setups for five manipulation tasks. Arrows illustrate the intended manipulations.}
    \label{fig:setting}
\end{wrapfigure}

\paragraph{Tasks and environments.}
We evaluate five real-world tasks (Figure~\ref{fig:setting}): \textit{Insert Flower}, \textit{Capsule Toy}, \textit{Insert Pen}, \textit{Make Tanghulu}, and \textit{Open Drawer}. These tasks cover grasping, precise insertion, and articulated-object interaction. Task, hardware, and data details are provided in Appendices~\ref{app:platform_tasks} and~\ref{app:data}.

\textbf{Evaluation metrics.}\enspace
We report three metrics. \textit{Task Completion Score} measures task execution quality on a normalized scale of $[0,10]$. \textit{Frame-level Macro-F1} compares per-frame predictions with ongoing, success, and failure annotations throughout the trajectory, assessing fine-grained status recognition. \textit{Trajectory-level F1} measures whether an annotated success or failure outcome is correctly recognized within a fixed window starting at the onset of the corresponding interval, assessing the timeliness and practical usability of outcome detection. Within each window, any failure prediction yields a failure decision; otherwise, any success prediction yields success, and detecting neither counts as a missed outcome. For each F1 metric, we average class-wise F1 over the ground-truth classes present in each rollout, then average across rollouts.

Unless otherwise specified, evaluations use 30 trials per model and task, thresholds of $0.5$ for the stop signal (SS) and failure signal (FS), and a 15-frame decision window (3\,s at a 0.2\,s inference interval). We report mean $\pm$ standard error (SE), with F1 in percentages. Scoring, annotation, and outcome-decision rules are detailed in Appendix~\ref{app:evaluation}.

\textbf{Models and training protocol.}\enspace
All experiments use $\pi_{0.5}$~\citep{intelligence2025pi_} as the policy backbone. Echo and the action-only baseline use the same successful and recovery demonstrations for action supervision, with matched training steps and optimizer settings. Echo additionally uses failure trajectories and pre-intervention failure segments for status supervision, excluding their actions from imitation.
Training details are provided in Appendix~\ref{app:implementation}.

\subsection{Main Results}
\label{sec:main_results}

%

\providecommand{\EchoNum}[2]{%
    \mbox{\ensuremath{#1{\scriptscriptstyle\,\pm\,#2}}}}
\providecommand{\EchoNumB}[2]{%
    \mbox{\boldmath\ensuremath{#1{\scriptscriptstyle\,\pm\,#2}}}}

\begin{table}[t]
    \centering
    \definecolor{EchoRowBlue}{RGB}{225,239,255}

    \begin{minipage}[t]{0.53\textwidth}
        \vspace{0pt}
        \caption{Main results on five real-world tasks.
        Score: Task Completion Score ($0$--$10$);
        Frame F1 and Traj. F1: frame-level and trajectory-level
        Macro-F1 (\%), respectively.
        Mean $\pm$ SE over 30 trials per model and task.}
        \label{tab:main_results}

        \begin{center}
        \scriptsize
        \setlength{\tabcolsep}{0pt}
        \renewcommand{\arraystretch}{1.15}
        \renewcommand{\tabularxcolumn}[1]{m{#1}}

        \begin{tabularx}{\linewidth}{
            @{}
            >{\raggedright\arraybackslash}m{0.21\linewidth}
            *{4}{>{\centering\arraybackslash}X}
            @{}
        }
            \toprule
            \multirow{2}{*}{\textbf{Task}}
            & \multicolumn{2}{c}{\textbf{Score}}
            & \multicolumn{2}{c}{\textbf{Status prediction of Echo}} \\
            \cmidrule(lr){2-3}\cmidrule(lr){4-5}
            & $\pi_{0.5}$ & \textbf{Echo}
            & \textbf{Frame F1} & \textbf{Traj. F1} \\
            \midrule

            Insert Flower
            & \EchoNum{5.67}{0.92}
            & \cellcolor{EchoRowBlue}\EchoNumB{8.33}{0.69}
            & \EchoNum{92.18}{0.71}
            & \EchoNum{96.67}{3.33} \\

            Capsule Toy
            & \EchoNum{8.40}{0.18}
            & \cellcolor{EchoRowBlue}\EchoNumB{8.57}{0.16}
            & \EchoNum{88.31}{0.57}
            & \EchoNum{100.00}{0.00} \\

            Insert Pen
            & \EchoNum{6.87}{0.84}
            & \cellcolor{EchoRowBlue}\EchoNumB{8.60}{0.46}
            & \EchoNum{89.33}{1.45}
            & \EchoNum{96.67}{3.33} \\

            Make Tanghulu
            & \EchoNum{4.33}{0.92}
            & \cellcolor{EchoRowBlue}\EchoNumB{5.33}{0.93}
            & \EchoNum{91.11}{1.13}
            & \EchoNum{93.33}{4.63} \\

            Open Drawer
            & \EchoNum{5.40}{0.77}
            & \cellcolor{EchoRowBlue}\EchoNumB{7.90}{0.26}
            & \EchoNum{88.01}{1.22}
            & \EchoNum{90.00}{5.57} \\

            \bottomrule
        \end{tabularx}
        \end{center}
    \end{minipage}%
    \hfill
    \begin{minipage}[t]{0.455\textwidth}
        \vspace{0pt}
        \caption{Comparison on Insert Flower.
        Score: $0$--$10$; F1: Macro-F1 (\%).
        Mean $\pm$ SE over 30 trials.
        $\dagger$: binary success/non-success evaluation;
        other F1 results use three-state evaluation.
        ``--'': not evaluated.}
        \label{tab:baseline_comparison}

        \begin{center}
        \scriptsize
        \setlength{\tabcolsep}{0pt}
        \renewcommand{\arraystretch}{1}
        \renewcommand{\tabularxcolumn}[1]{m{#1}}

        \begin{tabularx}{\linewidth}{
            @{}
            >{\raggedright\arraybackslash}m{0.25\linewidth}
            *{3}{>{\centering\arraybackslash}X}
            @{}
        }
            \toprule
            \multirow{2}{*}{\textbf{Method}}
            & \multirow{2}{*}{\textbf{Score}}
            & \multicolumn{2}{c}{\textbf{Status prediction}} \\
            \cmidrule(lr){3-4}
            & & \textbf{Frame F1} & \textbf{Traj. F1} \\
            \midrule


            CycleVLA
            & \EchoNum{3.00}{0.85}
            & \EchoNum{43.10}{1.80}
            & \EchoNum{36.67}{8.95} \\

            VLA-Corrector
            & \EchoNum{5.67}{0.92}
            & --
            & -- \\

            \rowcolor{EchoRowBlue}
            Echo
            & \EchoNumB{8.33}{0.69}
            & \EchoNumB{92.18}{0.71}
            & \EchoNumB{96.67}{3.33} \\

            \midrule

            Robometer$^\dagger$
            & --
            & \EchoNum{75.54}{7.05}
            & \EchoNum{90.00}{5.57} \\

            \rowcolor{EchoRowBlue}
            Echo$^\dagger$
            & \EchoNumB{8.33}{0.69}
            & \EchoNumB{93.09}{0.81}
            & \EchoNumB{96.67}{3.33} \\

            \bottomrule
        \end{tabularx}
        \end{center}
    \end{minipage}
\end{table}

\textbf{Task-status prediction.}\enspace
Table~\ref{tab:main_results} shows that Echo learns accurate task-status predictions without additional frame-level progress annotations. Strong frame-level performance indicates that the learned signals track execution states throughout the trajectory, while strong trajectory-level performance on most tasks supports their use for timely outcome recognition. Overall, the results support the proposed supervision scheme for learning signals that capture both local execution states and task outcomes.

\textbf{Task execution performance.}\enspace
Echo outperforms the action-only $\pi_{0.5}$ baseline on every task in Table~\ref{tab:main_results}. Since both models use the same action-supervision data, the consistent improvement supports jointly predicting task status and using the learned signals to condition action generation. The benefit therefore extends beyond monitoring execution to improving task completion. The analyses below examine when these gains are most pronounced and how the signals affect action generation.

\subsection{Comparison with Existing Methods}
\label{sec:signal_baseline}

We compare Echo with CycleVLA~\citep{ma2026cyclevla}, VLA-Corrector~\citep{pan2026vla}, and Robometer~\citep{liang2026robometer} on Insert Flower. CycleVLA augments action generation with stop and progress prediction. VLA-Corrector detects discrepancies between predicted and observed visual features to trigger corrective replanning. Robometer is an external reward model that estimates task progress and success from execution videos. Without a termination prediction, Robometer is evaluated using binary success/non-success Macro-F1; we also report Echo under this protocol. Implementations and adaptations to our evaluation protocol are detailed in Appendix~\ref{app:baselines}. Echo achieves the strongest status-prediction and task-completion performance on the applicable metrics (Table~\ref{tab:baseline_comparison}).

\subsection{Ablation and Mechanism Analysis}
\label{sec:ablations}

We next investigate why Echo improves execution and when its learned signals are most useful. We compare in-distribution and out-of-distribution conditions, ablate the signal--action architecture, directly intervene on the signals, and vary the amount of terminal-failure data.

\subsubsection{In-Domain and Out-of-Distribution Analysis}
\label{sec:ood_generalization}

We break down the main results for Insert Flower and Open Drawer into in-distribution (ID) and out-of-distribution (OOD) conditions, with 15 trials each (Table~\ref{tab:ood}). Details are in Appendix~\ref{app:platform_tasks}.

Echo's gains are larger under OOD conditions, while ID differences are modest and task-dependent (Table~\ref{tab:ood}). Status prediction remains broadly stable: frame-level performance changes modestly, and trajectory-level performance declines slightly under distribution shift.

Successful demonstrations emphasize nominal execution, whereas recovery demonstrations emphasize adjustment and correction. We hypothesize that status signals steer the policy toward the latter behaviors under distribution shift. The signal interventions below directly test this hypothesis.

\subsubsection{Component Ablations}
\label{sec:component_ablation}
\providecommand{\EchoNum}[2]{\mbox{\ensuremath{#1{\scriptscriptstyle\,\pm\,#2}}}}
\providecommand{\EchoNumB}[2]{\mbox{\boldmath\ensuremath{#1{\scriptscriptstyle\,\pm\,#2}}}}

\begin{table}[t]
    \centering
    \definecolor{EchoRowBlue}{RGB}{225,239,255}
    \begin{minipage}[t]{0.55\textwidth}
        \vspace{0pt}
        \caption{ID/OOD breakdown on Insert Flower and Open Drawer. Each condition contains 15 trials from the main evaluation. Score: $0$--$10$; F1: Echo's status metrics (\%). Results are mean $\pm$ SE.}
        \label{tab:ood}
        \begin{center}
        \scriptsize
        \setlength{\tabcolsep}{0pt}
        \renewcommand{\arraystretch}{1.15}
        \renewcommand{\tabularxcolumn}[1]{m{#1}}
        \begin{tabularx}{\linewidth}{@{}>{\raggedright\arraybackslash}m{0.22\linewidth}*{4}{>{\centering\arraybackslash}X}@{}}
            \toprule
            \multirow{2}{*}{\textbf{Metric / Model}}
            & \multicolumn{2}{c}{\textbf{Insert Flower}}
            & \multicolumn{2}{c}{\textbf{Open Drawer}} \\
            \cmidrule(lr){2-3}\cmidrule(lr){4-5}
            & \textbf{ID} & \textbf{OOD} & \textbf{ID} & \textbf{OOD} \\
            \midrule
            Score / $\pi_{0.5}$
            & \EchoNum{9.33}{0.67} & \EchoNum{2.00}{1.07}
            & \EchoNum{8.40}{0.13} & \EchoNum{2.40}{1.06} \\
            \rowcolor{EchoRowBlue}
            {Score / Echo}
            & \EchoNumB{8.67}{0.91} & \EchoNumB{8.00}{1.07}
            & \EchoNumB{8.93}{0.18} & \EchoNumB{6.87}{0.29} \\
            \midrule
            Frame F1 / Echo
            & \EchoNum{93.19}{0.93} & \EchoNum{91.17}{1.03}
            & \EchoNum{87.27}{1.02} & \EchoNum{88.74}{2.24} \\
            Traj. F1 / Echo
            & \EchoNum{100.00}{0.00} & \EchoNum{93.33}{6.67}
            & \EchoNum{93.33}{6.67} & \EchoNum{86.67}{9.09} \\
            \bottomrule
        \end{tabularx}
        \end{center}
    \end{minipage}%
    \hfill
    \begin{minipage}[t]{0.435\textwidth}
        \vspace{0pt}
        \caption{Component ablations on Insert Flower. $S\!\to\!A$ denotes action attention to both signal groups. Score: $0$--$10$; F1: \%. Mean $\pm$ SE over 30 trials per variant.}
        \label{tab:component_ablation}
        \begin{center}
        \scriptsize
        \setlength{\tabcolsep}{0pt}
        \renewcommand{\arraystretch}{1.15}
        \renewcommand{\tabularxcolumn}[1]{m{#1}}
        \begin{tabularx}{\linewidth}{@{}>{\raggedright\arraybackslash}m{0.23\linewidth}*{3}{>{\centering\arraybackslash}X}@{}}
            \toprule
            \textbf{Variant} & \textbf{Score} & \textbf{Frame F1} & \textbf{Traj. F1} \\
            \midrule
            \rowcolor{EchoRowBlue}
            {Echo}
            & \EchoNumB{8.33}{0.69} & \EchoNumB{92.18}{0.71} & \EchoNum{96.67}{3.33} \\
            \midrule
            w/o $S\!\to\!A$
            & \EchoNum{5.00}{0.93} & \EchoNum{90.78}{0.49} & \EchoNumB{100.00}{0.00} \\
            Full Attention
            & \EchoNum{7.67}{0.79} & \EchoNum{86.14}{1.40} & \EchoNum{100.00}{0.00} \\
            \midrule
            w/o Async.
            & \EchoNum{7.33}{0.82} & \EchoNum{87.29}{1.73} & \EchoNum{86.67}{6.31} \\
            Shared MLP
            & \EchoNum{6.67}{0.88} & \EchoNum{78.68}{3.31} & \EchoNum{83.33}{6.92} \\
            \bottomrule
        \end{tabularx}
        \end{center}
    \end{minipage}
\end{table}

We ablate signal--action attention, signal separation, and asynchronous denoising to examine how these choices contribute to status prediction and task execution. All ablations use Insert Flower, with each variant modifying one component while holding the remaining settings fixed (Table~\ref{tab:component_ablation}).

\textbf{Signal-to-action attention.}\enspace
Blocking action attention to both signal groups (\textit{w/o $S\!\to\!A$}) preserves signal quality but reduces task completion to near the action-only baseline. This indicates that the execution gains depend on actions conditioning on the learned signals.

\textbf{Signal separation.}\enspace
Allowing direct SS--FS attention (\textit{Full Attention}) reduces frame-level accuracy and task completion while preserving strong trajectory-level performance. We observe copying between the signals that weakens their distinct semantics. Sharing multilayer perceptrons (MLPs) across action, SS, and FS causes larger degradation in both prediction and execution, further supporting separate feature transformations. 

\textbf{Asynchronous denoising.}\enspace
Synchronous denoising uses the same flow time for actions and signals during training and inference. It causes only a modest decrease in frame-level accuracy but degrades task completion and timely outcome recognition. This supports making cleaner status estimates available earlier to guide subsequent action updates.

\subsubsection{Do Task-Status Signals Influence Action Generation?}
\label{sec:signal_intervention}

\begin{figure}[!t]
    \centering
    \begin{minipage}[t]{0.51\textwidth}
        \centering
        \vspace{0pt}
        \parbox[c][0.70\linewidth][c]{\linewidth}{%
            \includegraphics[pagebox=cropbox,width=\linewidth,height=0.70\linewidth,keepaspectratio]{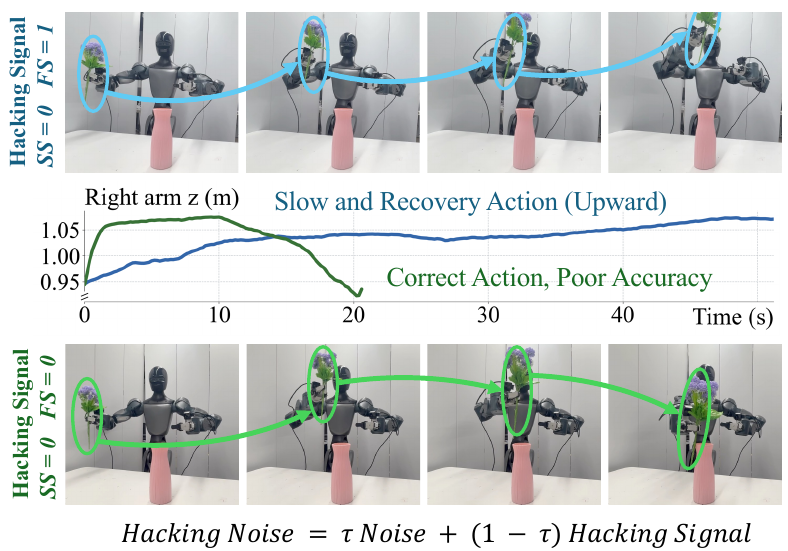}%
        }
        \caption{Signal interventions on Insert Flower. With $\text{SS}=0$, imposing $\text{FS}=1$ elicits upward recovery motions (blue), while $\text{FS}=0$ produces task-directed motions (green). Curves show right-arm height over time.}
        \label{fig:steering}
    \end{minipage}%
    \hfill
    \begin{minipage}[t]{0.47\textwidth}
        \centering
        \vspace{0pt}
        \parbox[c][0.70\linewidth][c]{\linewidth}{%
            \includegraphics[width=\linewidth,height=0.70\linewidth,keepaspectratio]{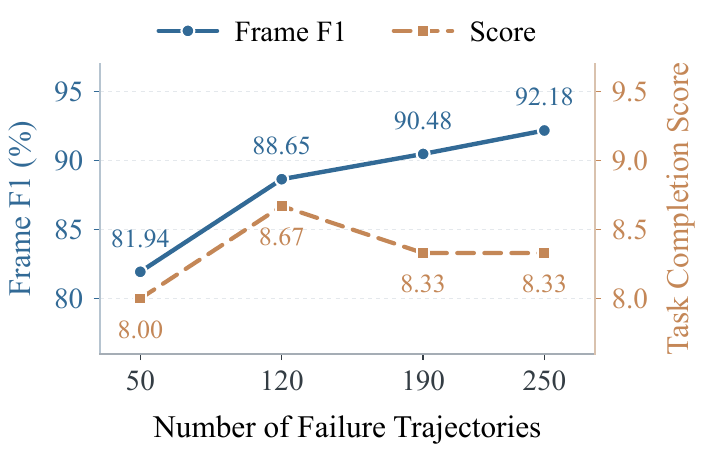}%
        }
        \caption{Effect of terminal-failure data quantity on Insert Flower, with its sampling proportion fixed. Curves show mean Frame-level Macro-F1 (left axis) and Task Completion Score (right axis) over 30 trials.}
        \label{fig:failure_data_scaling}
    \end{minipage}
\end{figure}

On Insert Flower, we impose prescribed signal targets at the corresponding noise level before each denoising step:
\begin{equation}
    \mathbf{x}^{q}_{\tau_q}
    \leftarrow
    (1-\tau_q)\mathbf{h}^{q}
    +\tau_q\boldsymbol{\epsilon}^{q},
    \qquad q\in\{S,F\},
    \label{eq:signal_intervention}
\end{equation}
Here, $\mathbf{h}^{q}\in\mathbb{R}^{H\times 1}$ is the prescribed target and $\boldsymbol{\epsilon}^{q}$ is Gaussian noise. We fix SS to zero and FS to either zero or one throughout the chunk, imposing normal execution or recoverable failure, respectively.

In Figure~\ref{fig:steering}, $\text{FS}=0$ produces task-directed motions, whereas sustained $\text{FS}=1$ elicits repeated upward recovery motions. This qualitative intervention shows that status inputs can directly steer action generation toward corrective behavior, supporting their use for biasing the policy toward adjustment and recovery when needed.

\subsubsection{Effect of Failure Data Quantity}
\label{sec:failure_data_scaling}

On Insert Flower, we vary the number of terminal-failure trajectories using nested subsets of 50, 120, 190, and 250 trajectories, with 250 used in the main experiments. The other three data types, training settings, and failure-data sampling proportion remain fixed. Thus, larger subsets provide more diverse failure examples without increasing their share of training samples.

Figure~\ref{fig:failure_data_scaling} shows that frame-level status prediction improves as the failure dataset grows, while task-completion scores remain broadly stable. This is consistent with the role of terminal-failure data: they supervise the distinction between terminal failure and successful completion, while their actions are excluded from imitation. Broader coverage of terminal-failure states thus enriches status supervision without directly supplying additional corrective actions.

The stable completion scores are also consistent with adjustment and recovery depending primarily on the pre-intervention and recovery data held fixed in this experiment. Pre-intervention failures supervise signals at recoverable errors, while recovery demonstrations provide corrective actions. Together, these observations support the complementary roles of terminal-failure supervision and recovery-related data. 

\subsection{Simulation Evaluation}
\label{sec:simulation}
 
We additionally evaluate a three-type Echo model on 150 fixed LIBERO-Recovery~\citep{liu2026libero} scenes spanning five LIBERO~\citep{liu2023libero} suites. Across 4,500 trials per model, Echo achieves 34.76\% task success, compared with 32.67\% for the action-only $\pi_{0.5}$ baseline trained on the same success and recovery data. Echo also achieves 89.97\% frame-level status accuracy. These results complement the real-world findings with a simulation evaluation of task completion and status prediction. Appendix~\ref{app:simulation} provides the data, training and evaluation protocols, per-seed results, and synchronized observations and signal traces.

\section{Conclusions}
We presented Echo, a compact framework that jointly denoises actions and task status signals within a single VLA. Echo learns from successful, failed, and recovery data without frame-level progress annotations, using controlled interactions and asynchronous denoising to guide action generation. Experiments on five real-world tasks demonstrate accurate status prediction and improved task completion, supporting the use of task status signals for both monitoring and action generation.

\subsection*{AI use statement}
We used generative AI tools to improve the clarity, grammar, and readability of the manuscript. All AI-assisted edits were reviewed by the authors, who take full responsibility for the final content of this work.



\bibliography{iclr2027_conference}
\bibliographystyle{iclr2027_conference}

\newpage
\appendix
\section{Appendix}

\providecommand{\EchoNA}{\textnormal{N/A}}

\subsection{Robot Platform and Task Details}
\label{app:platform_tasks}

\paragraph{Robot platform.}
Data collection and policy deployment use a wheeled humanoid robot with two seven-degree-of-freedom arms, gripper end effectors, and a waist--leg mechanism with six degrees of freedom. The mobile base remains stationary during task execution. The policy controls a total of 20 degrees of freedom through the two arm end effectors, the grippers, and the waist--leg mechanism. Visual observations come from three RGB cameras: one on the head and one on each wrist.

\paragraph{Common evaluation protocol.}
All tasks start from the same initial robot state. Task objects are placed at the evaluation positions described below. Each episode has a maximum duration of five minutes. An episode is classified as a failure if the task remains incomplete at the time limit, or if the robot reaches an unsafe posture indicating an imminent risk of tipping before completing the task.

\paragraph{Task objectives and evaluation positions.}
We evaluate five tasks: Insert Flower, Capsule Toy, Insert Pen, Make Tanghulu, and Open Drawer. Table~\ref{tab:app_task_details} summarizes their objectives and evaluation positions.

\begin{table}[h]
    \caption{Task objectives and evaluation positions. All tasks share the initial robot state and the five-minute episode limit. ID and OOD denote in-domain and out-of-distribution positions, respectively.}
    \label{tab:app_task_details}
    \begin{center}
    \small
    \renewcommand{\arraystretch}{1.15}
    \setlength{\tabcolsep}{6pt}
    \renewcommand{\tabularxcolumn}[1]{m{#1}}
    \begin{tabularx}{\linewidth}{@{}
        >{\raggedright\arraybackslash}m{0.18\linewidth}
        >{\raggedright\arraybackslash}m{0.42\linewidth}
        >{\raggedright\arraybackslash}X@{}}
        \toprule
        Task & Objective & Evaluation positions \\
        \midrule
        Insert Flower & Grasp a flower and insert it into a vase. & 3 random ID and 3 OOD tabletop positions; 5 trials each (30 total). \\
        \addlinespace[4pt]
        Capsule Toy & Rotate the machine's knob until a capsule is dispensed. & -- \\
        \addlinespace[4pt]
        Insert Pen & Grasp a pen and insert it into a pen holder. & 6 random tabletop positions; no explicit ID/OOD split (30 trials). \\
        \addlinespace[4pt]
        Make Tanghulu & Insert a bamboo skewer into a hawthorn fruit. & -- \\
        \addlinespace[4pt]
        Open Drawer & Pull the drawer open. & 1 ID and 1 OOD position; 15 trials each (30 total). \\
        \bottomrule
    \end{tabularx}
    \end{center}
\end{table}

\paragraph{Task outcomes and annotation boundaries.}
For Insert Flower, successful insertion constitutes task success. A dropped flower followed by the arm returning to its reset position without attempting to grasp it again constitutes failure. Insert Pen follows the analogous insertion and failure protocol.

For Make Tanghulu, success requires inserting the bamboo skewer into the hawthorn and retaining the fruit on the skewer after release. An episode is classified as a failure if insertion is not achieved or if the hawthorn falls off after release. The latter receives partial credit of 5 points, as specified in Table~\ref{tab:app_scoring}.
Open Drawer succeeds when the drawer is pulled open, and Capsule Toy succeeds when knob rotation causes a capsule to be dispensed. Episodes that remain unsuccessful are subject to the common termination conditions above.

\subsection{Evaluation Protocol and Task Scoring}
\label{app:evaluation}

\subsubsection{Task Completion Score}
\label{app:scoring}

Each episode has a maximum task score of 10. Successful episodes receive 10 points before task-specific deductions. Failed episodes receive 0 points, except in Make Tanghulu when insertion is achieved but the hawthorn subsequently falls off the skewer after release; these failed episodes receive 5 points as partial credit. For Insert Flower and Insert Pen, each contact event incurs a one-point deduction, capped at three points. For Capsule Toy and Open Drawer, each empty grasp attempt incurs a one-point deduction, capped at five points.


\begin{table}[t]
    \caption{Task-specific scoring rules. Successful episodes receive 10 points minus the applicable deductions. Failed episodes receive 0 points, except for Make Tanghulu episodes in which the hawthorn falls off after insertion and release; these receive 5 points as partial credit.}
    \label{tab:app_scoring}
    \begin{center}
    \small
    \renewcommand{\arraystretch}{1.2}
    \setlength{\tabcolsep}{8pt}
    \renewcommand{\tabularxcolumn}[1]{m{#1}}
    \begin{tabularx}{\linewidth}{@{}
        >{\raggedright\arraybackslash}m{0.19\linewidth}
        >{\raggedright\arraybackslash}X
        >{\centering\arraybackslash}m{0.16\linewidth}@{}}
        \toprule
        Task & Deduction rule & Maximum deduction \\
        \midrule
        Insert Flower & 1 point per contact with the vase wall. & 3 points \\
        \addlinespace[5pt]
        Capsule Toy & 1 point per empty grasp attempt. & 5 points \\
        \addlinespace[5pt]
        Insert Pen & 1 point per contact with the pen holder. & 3 points \\
        \addlinespace[5pt]
        Make Tanghulu & 5 points if the hawthorn falls off the skewer after release. & 5 points \\
        \addlinespace[5pt]
        Open Drawer & 1 point per empty grasp attempt. & 5 points \\
        \bottomrule
    \end{tabularx}
    \end{center}
\end{table}

\subsubsection{Frame-Level and Trajectory-Level Metrics}
\label{app:metrics}

Predicted execution states follow the SS/FS threshold rule defined in the main text. We compute each metric separately for each episode and then report its mean and standard error across episodes.

\paragraph{Frame-level Macro-F1.}
For episode $i$ and class $c$, we compute
\begin{equation}
    F1_{i,c} = \frac{2TP_{i,c}}{2TP_{i,c}+FP_{i,c}+FN_{i,c}},
    \qquad
    F^{\mathrm{frame}}_i
    =\frac{1}{|\mathcal C_i|}\sum_{c\in\mathcal C_i}F1_{i,c},
\end{equation}
where $\mathcal C_i$ contains the execution-state classes present in the ground-truth annotations of episode $i$.

\paragraph{Trajectory outcome recognition.}
For each annotated success or failure interval, the evaluator examines the first $W$ available predictions starting at its ground-truth boundary. If any prediction is failure, the predicted outcome is failure. Otherwise, it is success if any prediction is success, and ongoing if neither is detected. When the annotated interval contains fewer than $W$ frames, all available frames in that interval are used.

We compute outcome F1 within each episode over the outcome classes present in its ground truth. An ongoing prediction counts as a missed outcome. The resulting per-episode values are then aggregated across episodes.
For the results reported in the main text, we use $W=15$ frames. The policy is queried at 5 frames per second, so this window corresponds to 3 seconds.

\paragraph{Aggregation across episodes.}
Each model is evaluated for 30 episodes per task. For an episode-level metric $q_i$, we report
\begin{equation}
    \bar q=\frac{1}{N}\sum_{i=1}^{N}q_i,
    \qquad
    \mathrm{SE}(\bar q)
    =\sqrt{\frac{\sum_{i=1}^{N}(q_i-\bar q)^2}{N(N-1)}}.
\end{equation}
The same episode-first aggregation is used for task score, frame-level Macro-F1 and trajectory-level F1.

\subsection{Data Sampling Ratio}
\label{app:data}

For every task, we use the same training sampling ratio of $20:1:2:20$ for Success, Failure, Pre-intervention failure, and Recovery data, respectively.

\subsection{Implementation, Hyperparameters, and Inference Cost}
\label{app:implementation}

\subsubsection{Training and Deployment Settings}

The main experiments and most ablations use $\pi_{0.5}$. For each task, Echo and its action-only baseline use matched optimizer settings and training steps. Actions, SS and FS use separate token groups with separate feature-transformation MLPs and output projections. Signal flow times follow the asynchronous schedule in the method section, with $\tau_{\text{clean}}=0.5$ shared by SS and FS. Both signals reach their clean endpoints when the action flow time reaches $\tau_A=0.5$. Table~\ref{tab:app_hparams} summarizes the training and inference hyperparameters.

\begin{table}[t]
    \caption{Training, model and deployment hyperparameters.}
    \label{tab:app_hparams}
    \begin{center}
    \small
    \setlength{\tabcolsep}{8pt}
    \renewcommand{\arraystretch}{1.1}
    \begin{tabular}{@{}llr@{}}
        \toprule
        \textbf{Hyperparameter}
        & \textbf{Description}
        & \textbf{Value} \\
        \midrule
        $H$ & Action chunk length & 60 \\
        $\lambda_A$ & Action loss weight & 1 \\
        $\lambda_S$ & Stop signal loss weight (non-padding chunks) & 0.1 \\
        $\lambda_F$ & Failure signal loss weight (non-padding chunks) & 1 \\
        $\tau_{\text{clean}}$
        & Signal clean-endpoint time (SS and FS) & 0.5 \\
        $\delta_s$ & Stop signal threshold & 0.5 \\
        $\delta_f$ & Failure signal threshold & 0.5 \\
        $W$ & Trajectory outcome window & 15 frames (3 s) \\
        Query rate & Policy inference frequency & 5 Hz \\
        \bottomrule
    \end{tabular}
    \end{center}
\end{table}

For chunks containing padding, both signal loss weights are set to 1, following the supervision rule in the main text.

\subsubsection{Inference Overhead}
\label{app:latency}

Despite increasing the generated-token sequence from 60 to 180 tokens and adding separate MLPs, Echo increases inference latency on an NVIDIA RTX 4090 by only 5 ms, from 115 ms to 120 ms, introducing little additional inference overhead.

\subsection{Implementation of the Motivating Experiment}
\label{app:motivating_experiment}

The motivating experiment examines how intermediate action and status estimates evolve during a synchronous denoising process. Its purpose is to determine when status information becomes available before introducing the asynchronous schedule.

\paragraph{Synchronous reference model.}
We use a baseline that denoises actions, SS, and FS synchronously, assigning the same flow time to all three groups:
\begin{equation}
    \tau_A=\tau_S=\tau_F=\tau.
\end{equation}
This corresponds to $\tau_{\text{clean}}=0$ during both training and inference, instead of the $\tau_{\text{clean}}=0.5$ used by Echo. All other architectural designs are the same as Echo, including the signal--action attention mask, separate MLPs, and group-specific projections. In particular, action tokens and each signal group can attend to one another, while direct attention between SS and FS remains blocked. The reference model jointly predicts actions and both signals; it is distinct from the action-only $\pi_{0.5}$ baseline.

\paragraph{Intermediate clean-sample estimates.}
We record all ten denoising steps, indexed from 0 to 9, within a single inference query. The observation, robot state, and instruction are held fixed throughout that query. At each step $k$, we record the current noisy state $\mathbf{x}_{\tau_k}$ and the velocity predicted by the model. For each group $q\in\{A,S,F\}$, we reconstruct its clean-sample estimate as
\begin{equation}
    \hat{\mathbf{x}}^{q}_{0,k}
    =
    \mathbf{x}^{q}_{\tau_k}
    -
    \tau_k\mathbf{v}^{q}_{\theta}
    (\mathbf{x}_{\tau_k},\tau_k,\mathbf{c}).
    \label{eq:app_intermediate_estimate}
\end{equation}
The velocity prediction depends on the joint noisy input and the conditioning context $\mathbf{c}$. This reconstruction follows from the linear flow path: if the predicted velocity equals $\boldsymbol{\epsilon}^{q}-\mathbf{x}^{q}_0$, subtracting $\tau_k$ times that velocity recovers $\mathbf{x}^{q}_0$. With a learned velocity, it provides an estimate of the clean output at the current step.

The reconstruction is used for analysis and does not replace the noisy state in the inference procedure. This distinction is important: directly plotting $\mathbf{x}_{\tau_k}$ would mix the predicted content with the noise remaining at that flow time. The experiment instead compares clean-sample estimates produced at different denoising steps.

\paragraph{Action and signal visualizations.}
Figure~\ref{fig:speed}(a) plots the right-arm vertical action component across the $H=60$ chunk positions, expressed in millimeters. Each curve corresponds to a denoising step, with darker curves indicating later steps. Figure~\ref{fig:speed}(b) summarizes each reconstructed signal chunk using its mean:
\begin{equation}
    \hat{\bar{s}}_k
    =\frac{1}{H}\sum_{j=0}^{H-1}\hat{s}_{0,k,j},
    \qquad
    \hat{\bar{f}}_k
    =\frac{1}{H}\sum_{j=0}^{H-1}\hat{f}_{0,k,j}.
    \label{eq:app_intermediate_signal_means}
\end{equation}
The plotted steps are indexed from 0 to 9. SS and FS use separate vertical axes, both marked with a $10^{-2}$ scale, so their displayed magnitudes should be read from the corresponding axes.

\paragraph{Interpretation.}
In the plotted example, SS is close to zero from the earliest step, and FS approaches its later estimate within the first few steps. The action curves continue to change, including between the last two plotted steps. The figure therefore illustrates that useful status estimates can stabilize earlier than the continuous action sequence. It motivates making cleaner signal estimates available to subsequent action updates; it does not imply that every signal prediction is correct or that all action dimensions require the same number of denoising steps.

\subsection{Baseline Implementations and Performance Analysis}
\label{app:baselines}

We compare action-only $\pi_{0.5}$, CycleVLA, VLA-Corrector, Robometer and Echo on Insert Flower. This section distinguishes original methods from adaptations in our setup.

\paragraph{Action-only $\pi_{0.5}$.}
The baseline uses the same successful and recovery action-supervision data as Echo, with matched optimizer settings and training steps. It predicts actions without SS or FS, so status metrics are not assigned to it. During implementation, we ensured that key settings including training data, model learning rate, and optimizer were consistent with Echo. The model was trained for the same number of steps until convergence.

\begin{figure}[t]
    \centering
        \centering
        \vspace{0pt}
\includegraphics[pagebox=cropbox,width=\linewidth]{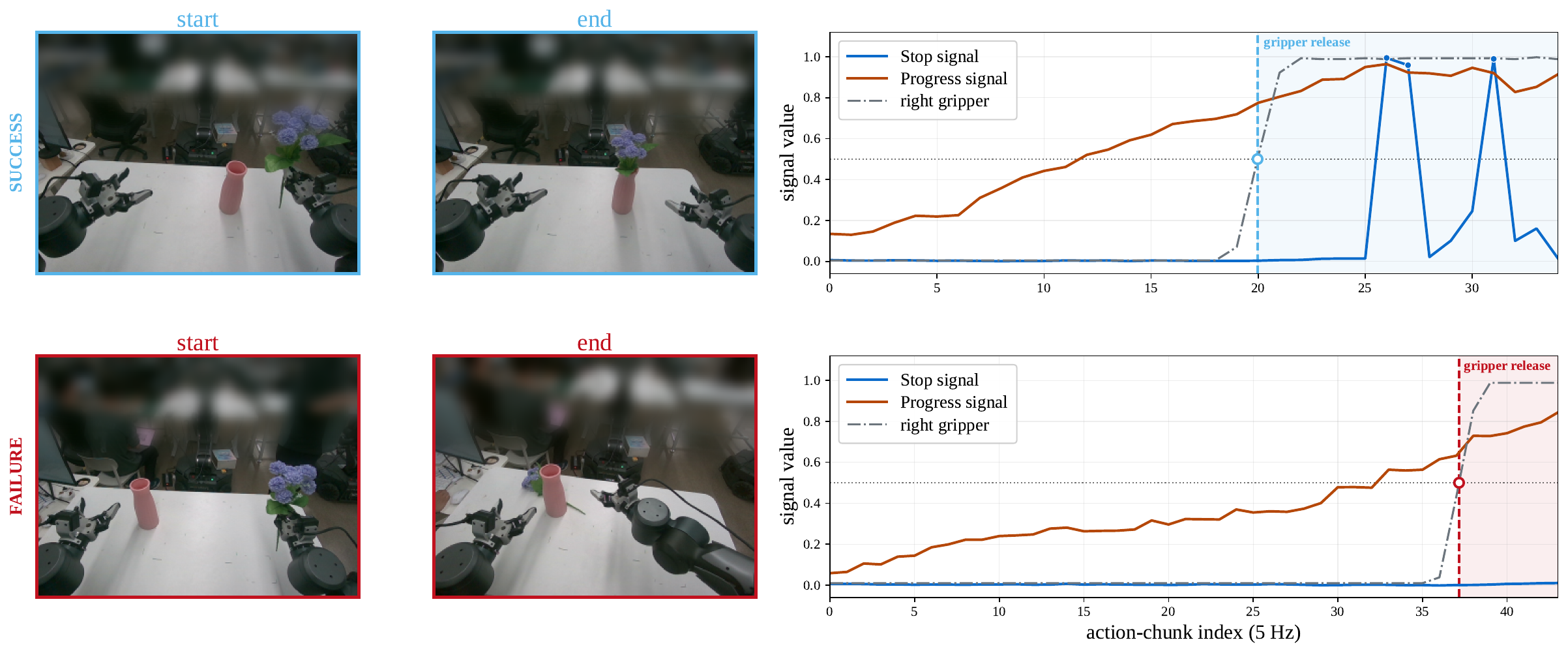}%
        \caption{Signal values estimated by CycleVLA. The model exhibits weak discriminative capability for real-world failures.
        }
        \label{fig:cycleVLA}
\end{figure}

\paragraph{CycleVLA.}
CycleVLA combines a progress-aware VLA with external failure prediction and subtask backtracking. 
When reproducing its performance, we trained using demonstration success data and DAgger recovery data. And we adopted a learning rate of 5e-5 and the Adam optimizer for 20k training steps (aligned with the training protocol of Echo). Since its stop signal is only supervised on the final frame, and success/failure judgment mainly relies on the progress signal, we take the maximum stop signal within the model’s actually executed frames and the average of the corresponding progress signals, apply a threshold of 0.5 to both quantities, and construct frame-level and trajectory-level state prediction discrimination accordingly. As illustrated in Figure \ref{fig:cycleVLA}, since CycleVLA does not leverage real-world policy failure data, it exhibits weak discriminative capability for real-world failures.

\paragraph{VLA-Corrector.}
The VLA-Corrector considered here compares predicted and observed visual feature evolution using a latent-space vision monitor. Persistent deviations trigger action-chunk truncation and corrective replanning with Online Gradient Guidance. It does not itself establish terminal failure, so status metrics are not assigned to it.

When reproducing its performance, we adopted the frozen baseline $\pi_{0.5}$ model and trained the corrector for 10k iterations following the recommended configuration in its original code. In practical testing, we corrected the VLA action chunk via the Online Gradient Guidance approach and the cutoff-only method, then we compared their performance and reported the better-performing one.

\paragraph{Robometer.}
Robometer is an external video-language reward model with progress and success predictions trained using trajectory-based supervision, including preference comparisons. 
When reproducing its performance, we trained the officially released Robometer-4B base model via LoRA fine-tuning. We trained for 20k iterations on all data of the Insert Flower task using its officially recommended configuration and evaluated its performance offline. Since Robometer does not provide estimations for task termination, we classify all ``in-progress'' samples as failures during evaluation, reducing the original 3-class classification problem to binary classification. We compute the final metrics by applying a threshold of 0.5 on the success probability output by the model.

\begin{figure}[t]
    \centering
        \centering
        \vspace{0pt}
\includegraphics[pagebox=cropbox,width=\linewidth]{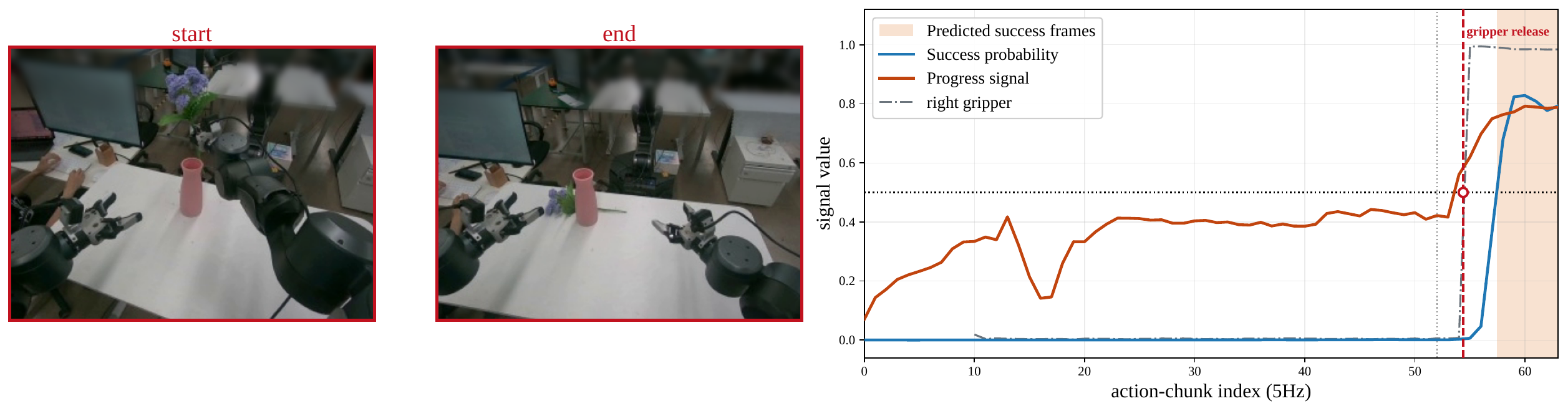}%
        \caption{Signal values estimated by Robometer. The model sometimes lacks discriminative capability for real-world failures.
        }
        \label{fig:Robometer}
\end{figure}

As shown in Figure \ref{fig:Robometer}, similar to CycleVLA, Robometer sometimes lacks discriminative ability for real-world failures.

\subsection{Detailed Task-Status Prediction Results}
\label{app:status_details}

Table~\ref{tab:app_status_details} provides the complete numerical results for Echo under the default evaluation settings: $\delta_s=\delta_f=0.5$ and $W=15$ frames. In addition to the frame-level and trajectory-level F1 scores reported in the main text, we provide overall frame accuracy, frame-level recall for each execution state, and trajectory-level success and failure recall.

\begin{figure}[t]
    \centering
    \includegraphics[width=\linewidth]{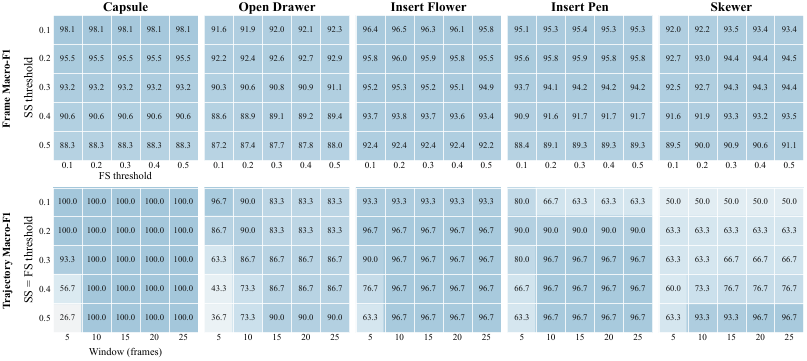}
    \caption{Hyperparameter sensitivity across five tasks. Top: frame-level Macro-F1 under independently varied SS and FS thresholds. Bottom: trajectory-level F1 under a shared SS/FS threshold and varying outcome-window lengths. Values are percentages. Capsule and Skewer denote Capsule Toy and Make Tanghulu, respectively. The main results use $\delta_s=\delta_f=0.5$ and $W=15$ frames.}
    \label{fig:app_heatmap}
\end{figure}

\providecommand{\EchoDetailNum}[2]{\mbox{\ensuremath{#1{\scriptscriptstyle\,\pm\,#2}}}}
\providecommand{\EchoDetailRecall}[2]{\mbox{\ensuremath{#1}\,{\scriptsize (#2)}}}

\begin{table}[t]
    \caption{Detailed task-status prediction results for Echo on five tasks.
    All metrics are percentages.
    Frame F1 denotes frame-level Macro-F1 and Traj.\ F1 denotes trajectory-level F1;
    both are reported as mean $\pm$ SE over 30 episodes per task.
    Trajectory recall includes the number of correctly recognized outcomes over the number of ground-truth outcomes in parentheses.
    A dash indicates that no ground-truth failure examples are present.}
    \label{tab:app_status_details}
    \centering
    \small
    \setlength{\tabcolsep}{3pt}
    \renewcommand{\arraystretch}{1}
    \renewcommand{\tabularxcolumn}[1]{m{#1}}

    \begin{tabularx}{\linewidth}{@{}
        >{\raggedright\arraybackslash}m{0.20\linewidth}
        *{3}{>{\centering\arraybackslash}X}@{}}
        \toprule
        \multicolumn{4}{@{}l}{\textbf{(a) Overall performance}} \\
        \addlinespace[3pt]
        \textbf{Task} & \textbf{Frame F1} & \textbf{Traj.\ F1} & \textbf{Frame Acc.} \\
        \midrule
        Insert Flower & \EchoDetailNum{92.18}{0.71} & \EchoDetailNum{96.67}{3.33} & 96.00 \\
        \addlinespace[2pt]
        Capsule Toy & \EchoDetailNum{88.31}{0.57} & \EchoDetailNum{100.00}{0.00} & 94.58 \\
        \addlinespace[2pt]
        Insert Pen & \EchoDetailNum{89.33}{1.45} & \EchoDetailNum{96.67}{3.33} & 92.18 \\
        \addlinespace[2pt]
        Make Tanghulu & \EchoDetailNum{91.11}{1.13} & \EchoDetailNum{93.33}{4.63} & 95.67 \\
        \addlinespace[2pt]
        Open Drawer & \EchoDetailNum{88.01}{1.22} & \EchoDetailNum{90.00}{5.57} & 96.51 \\
        \bottomrule
    \end{tabularx}

    \vspace{8pt}

    \begin{tabularx}{\linewidth}{@{}
        >{\raggedright\arraybackslash}m{0.20\linewidth}
        *{3}{>{\hsize=0.8\hsize\linewidth=\hsize\centering\arraybackslash}X}
        *{2}{>{\hsize=1.3\hsize\linewidth=\hsize\centering\arraybackslash}X}@{}}
        \toprule
        \multicolumn{6}{@{}l}{\textbf{(b) Class-wise recognition}} \\
        \addlinespace[3pt]
        {}
        & \multicolumn{3}{c}{\textbf{Frame-level recall}}
        & \multicolumn{2}{c}{\textbf{Trajectory-level recall}} \\
        \cmidrule(lr){2-4}\cmidrule(lr){5-6}
        \textbf{Task} & Ongoing & Success & Failure & Success & Failure \\
        \midrule
        Insert Flower & 99.96 & 77.25 & 79.78
        & \EchoDetailRecall{96.00}{24/25} & \EchoDetailRecall{100.00}{5/5} \\
        \addlinespace[3pt]
        Capsule Toy & 100.00 & 68.02 & --
        & \EchoDetailRecall{100.00}{30/30} & -- \\
        \addlinespace[3pt]
        Insert Pen & 100.00 & 77.17 & 75.00
        & \EchoDetailRecall{96.43}{27/28} & \EchoDetailRecall{100.00}{2/2} \\
        \addlinespace[3pt]
        Make Tanghulu & 99.96 & 78.60 & 77.28
        & \EchoDetailRecall{93.75}{15/16} & \EchoDetailRecall{92.86}{13/14} \\
        \addlinespace[3pt]
        Open Drawer & 99.79 & 66.06 & --
        & \EchoDetailRecall{90.00}{27/30} & -- \\
        \bottomrule
    \end{tabularx}
\end{table}

Ongoing-state recall exceeds 99\% on every task, while frame-level recognition of success and failure is lower. The trajectory-level results separately show whether each outcome is recognized within the evaluation window. The counts accompanying trajectory recall make the different numbers of successful and failed episodes explicit.

\subsection{Hyperparameter Sensitivity}
\label{app:sensitivity}

Figure~\ref{fig:app_heatmap} evaluates status-prediction sensitivity across all five tasks using two parameter grids. For frame-level evaluation, we independently vary the SS and FS thresholds over $\{0.1,0.2,0.3,0.4,0.5\}$. For trajectory-level evaluation, we use a shared threshold $\delta_s=\delta_f=\delta$ with $\delta\in\{0.1,0.2,0.3,0.4,0.5\}$ and vary the outcome window over $W\in\{5,10,15,20,25\}$ frames. At 5 Hz, these windows correspond to 1--5 seconds. Each setting is evaluated on the recorded episodes, with F1 computed per episode and then averaged across episodes.

\paragraph{Frame-level threshold sensitivity.}
Frame-level performance is generally more sensitive to the SS threshold than to the FS threshold over the tested ranges. For example, Capsule Toy changes from 98.1 at an SS threshold of 0.1 to 88.3 at 0.5, while remaining unchanged across the tested FS thresholds. The other tasks also show relatively modest changes along the FS-threshold axis, although the best SS threshold varies across tasks. Thus, a single threshold setting does not maximize frame-level F1 for every task.

\paragraph{Trajectory-level threshold and window sensitivity.}
Short outcome windows can miss predictions that become correct later in the outcome interval. At the default shared threshold of 0.5, increasing $W$ from 5 to 15 frames raises trajectory-level F1 from 26.7 to 100.0 on Capsule Toy, from 36.7 to 90.0 on Open Drawer, and from 63.3 to 96.7 on both Insert Flower and Insert Pen. Make Tanghulu increases from 63.3 to 93.3 over the same range and reaches 96.7 with a 20-frame window.

Longer windows do not compensate for every threshold choice. For example, Make Tanghulu remains at 50.0 across all tested window lengths when the shared threshold is 0.1. The heatmaps therefore show a task-dependent interaction between thresholds and the time allowed for outcome recognition. Around the default setting, several tasks exhibit a plateau once the window is sufficiently long, whereas Make Tanghulu benefits from a slightly longer window. These results support reporting the common default configuration together with its sensitivity, rather than assuming that larger windows or lower thresholds always improve recognition.

The window $W$ is used only for trajectory-level evaluation and does not change the frame-level predictions. These offline comparisons characterize status evaluation; they do not alter the actions or task-completion scores of the recorded episodes.

\subsection{Compatibility with External Monitoring and Planning}
\label{app:external}

Echo improves the VLA policy itself by allowing learned task-status information to participate in action generation. This objective is compatible with external monitoring, recovery and planning modules. SS and FS can also provide execution-status information to a higher-level controller using the signal semantics defined in the main text.

An external controller could combine these signals with broader temporal context or additional observations to verify completion, request intervention or replan. Echo supplies policy-internal status estimates and corrective action generation, while the external module adds its own decision logic. This is architectural compatibility; improved performance from a particular combined system requires a separate evaluation.

\subsection{Simulation Experiments}
\label{app:simulation}

\begin{table}[t]
    \centering
    \caption{Simulation results on the fixed LIBERO-Recovery scene set. Success rates and frame-level signal accuracy are percentages. Each seed has 1,500 trials per model. Average uses equal seed weights for task success and evaluated-frame counts as weights for signal accuracy. Echo uses $\tau_{\mathrm{clean}}=0.7$.}
    \label{tab:simulation}
    \small
    \setlength{\tabcolsep}{7pt}
    \renewcommand{\arraystretch}{1.12}
    \begin{tabular}{@{}lrrrr@{}}
        \toprule
        \multirow{2}{*}{\textbf{Evaluation seed}}
        & \multirow{2}{*}{\textbf{Trials/model}}
        & \multicolumn{2}{c}{\textbf{Task success}}
        & \textbf{Signal accuracy} \\
        \cmidrule(lr){3-4}
        & & $\pi_{0.5}$ & Echo & Echo \\
        \midrule
        7 & 1,500 & 32.87 & 34.47 & 89.76 \\
        17 & 1,500 & 31.07 & 34.67 & 90.55 \\
        27 & 1,500 & 34.07 & 35.13 & 89.58 \\
        \midrule
        \textbf{Average} & \textbf{4,500 total} & \textbf{32.67} & \textbf{34.76} & \textbf{89.97} \\
        \bottomrule
    \end{tabular}
\end{table}

We complement the real-world experiments with a three-type simulation study based on LIBERO~\citep{liu2023libero} and LIBERO-Recovery. This study evaluates both task completion from recovery states and task-status prediction using simulator success checks.
 
\paragraph{Environment and evaluation scenes.}
We use LIBERO-Recovery's original 880-scene asset release and fix 150 scenes: 30 each from the spatial, object, goal, LIBERO-10, and LIBERO-90 suites. Each scene provides a task definition and a recorded simulator state. The policy observes third-person and wrist RGB images and the robot state, and produces seven-dimensional end-effector and gripper actions. We use 20\,Hz control, execute eight actions per policy query, and allow 520 actions (26\,s) per rollout. For each scene, we evaluate five trials with an initially open gripper and five with a closed gripper, applying independent uniform object-position perturbations of $\pm2$\,cm in the horizontal plane. Seeds 7, 17, and 27 retain the same scenes and vary trial randomness, yielding 4,500 trials per model. Both models use the same frozen trial specifications and task instructions.

\begin{figure}[t]
    \centering
    \includegraphics[width=\textwidth]{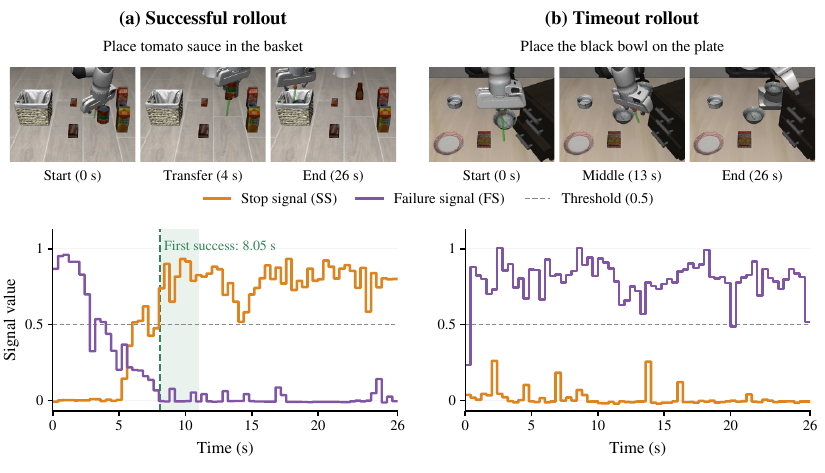}
    \caption{Task status signals in simulation. Each column shows three observations, cropped to the task region, above the stop signal (SS, orange) and failure signal (FS, purple), averaged over all $H=50$ chunk predictions. Left: successful tomato-sauce placement; the green line marks first success at 8.05\,s, followed by the shaded 3-second scoring window. Right: bowl placement reaching the 26-second timeout, with SS remaining below threshold. Both traces cover all 520 actions, including after success. Predictions are held between queries without smoothing or clipping; final images are recorded after action 520.}
    \label{fig:simulation_signals}
\end{figure}

\paragraph{Training data and configuration.}
The training pool contains 6,193 successful demonstrations from the public LIBERO datasets, 3,184 recovery demonstrations from the LIBERO-Recovery expert dataset, and 9,521 unsuccessful $\pi_{0.5}$ rollouts collected from LIBERO-Recovery scenes. The latter reach the 520-action limit without task success and provide failure supervision. Echo uses these three data types, without a separate pre-intervention category; the action-only $\pi_{0.5}$ baseline uses success and recovery data. Failure actions are excluded from imitation. Sampling weights are success:failure:recovery $=40:1:20$ for Echo and success:recovery $=2:1$ for the baseline. Both models start from the same released LIBERO-tuned $\pi_{0.5}$ checkpoint and are fine-tuned for 2,500 updates with training seed 42, global batch size 128 on four GPUs, and AdamW with 125 warmup updates followed by cosine learning-rate decay from $10^{-5}$ to $10^{-6}$. Echo uses $H=50$ and $\tau_{\mathrm{clean}}=0.7$.
 
SS targets are zero at valid positions and one at padded positions for all three data types, including recovery, with supervision over the full chunk.
 
\paragraph{Task success and signal evaluation.}
Task success is counted if the simulator reports success at least once within 520 actions. The baseline terminates at its first success; Echo continues to the full horizon so that signals can be inspected afterward. Signals do not terminate Echo's rollout. A timeout counts as unsuccessful task execution, but is not treated as a ground-truth terminal-failure state for signal evaluation: the simulator does not provide a terminal-failure onset annotation.
 
We report frame-level status accuracy using whole-chunk means from Eq.~\ref{eq:signal_average}, with both thresholds set to 0.5. Before the first success and throughout timeout rollouts, the target is Ongoing and a prediction is correct when $\bar{s}_t<0.5$. From the first success through the following 3\,s, the target is Success and a prediction is correct when $\bar{s}_t>0.5$ and $\bar{f}_t<0.5$; subsequent frames are excluded. Values are held between policy queries and only recorded observation frames are scored. Windows extending past the horizon are truncated. Overall accuracy is the total correct-frame count divided by the total evaluated-frame count, rather than the real-world Macro-F1 metric. This evaluates ongoing/success recognition; it does not separately measure the timing of terminal failures.
 
\paragraph{Results.}
Table~\ref{tab:simulation} reports the complete results. Echo succeeds on 1,564 of 4,500 trials (34.76\%), compared with 1,470 (32.67\%) for the action-only baseline, an increase of 2.09 percentage points. Echo is higher on all three evaluation seeds. Its frame-level signal accuracy is 89.97\% (1,725,710/1,918,181 evaluated frames). The three evaluation seeds share one trained checkpoint per model and are not independent training replicates. The default simulator renderer does not guarantee pixel-identical images across repeated runs with fixed seeds.

Figure~\ref{fig:simulation_signals} shows two illustrative rollouts. In the successful rollout, SS rises around task completion while FS decreases. In the timeout rollout, SS remains below threshold while FS varies during continued attempts. The complete traces also show behavior after the successful scoring window; these later frames do not contribute to signal accuracy.

\end{document}